\documentclass[a4paper, 10pt, conference]{ieeeconf}      
\usepackage{FG2021}

\FGfinalcopy 

\IEEEoverridecommandlockouts                              
\usepackage[accsupp]{axessibility} 
\usepackage{times}
\usepackage{epsfig}
\usepackage{graphicx}
\usepackage{amsmath}
\usepackage{amssymb}
\usepackage{makecell}
\let\labelindent\relax
\usepackage{enumitem}

\usepackage{multirow}
\usepackage{xcolor}

\usepackage{color, colortbl}
\definecolor{LightCyan}{rgb}{0.88,1,1}
\newcommand{\RNum}[1]{\uppercase\expandafter{\romannumeral #1\relax}}
\newcommand{\cmmnt}[1]{}
\definecolor{myblue1}{RGB}{9, 50, 200}

\newcommand{\etal}{\textit{et al.}}

\def\FGPaperID{131} 

\title{\LARGE \bf
	Heterogeneous Face Frontalization via Domain Agnostic Learning
}

\author{\parbox{16cm}{\centering
		{\large Xing Di$^1$ and Shuowen Hu$^2$ and Vishal M Patel$^1$}\\
		{\normalsize
			$^1$ Department of Electrical and Computer Engineering, Johns Hopkins University, Maryland, USA\\
			$^2$  U.S. Army DEVCOM Army Research Laboratory, Maryland, USA }}
	\thanks{This work was supported by ARO grant W911NF2110135.}
}

\begin{document}
%
%
%




\IEEEoverridecommandlockouts\pubid{\makebox[\columnwidth]{978-1-6654-3176-7/21/\$31.00~\copyright{}2021 IEEE \hfill}
	\hspace{\columnsep}\makebox[\columnwidth]{ }}

\ifFGfinal
\thispagestyle{empty}
\pagestyle{empty}
\else
\author{Anonymous FG2021 submission\\ Paper ID \FGPaperID \\}
\pagestyle{plain}
\fi

\maketitle


\begin{abstract}
	Recent  advances  in  deep  convolutional  neural  networks  (DCNNs) have shown impressive performance improvements on thermal to visible face synthesis and matching problems. However, current DCNN-based synthesis models do not perform well on thermal faces with large pose variations.    In order to deal with this problem, heterogeneous face frontalization methods are needed in which a model takes a thermal profile face image and generates a frontal visible face.  This is an extremely difficult problem due to the large domain as well as large pose discrepancies between the two modalities. 
	Despite its applications in biometrics and surveillance, this problem is relatively unexplored in the literature.  We propose a domain agnostic learning-based generative adversarial network (DAL-GAN) which can synthesize   frontal views in the visible domain from thermal faces with pose variations. 	DAL-GAN consists of a generator with  an auxiliary classifier and two discriminators which capture both local and global texture discriminations for better  synthesis.  A contrastive constraint is enforced in the latent space of the generator with the help of a dual-path training strategy, which improves the feature vector's discrimination.  Finally, a multi-purpose loss function is utilized to guide the network in synthesizing identity-preserving cross-domain frontalization.   Extensive experimental results demonstrate that DAL-GAN can generate better quality frontal views compared to the other baseline methods. 
\end{abstract}

\section{Introduction}

\begin{figure}
	\centering
	\includegraphics[width=0.9\linewidth]{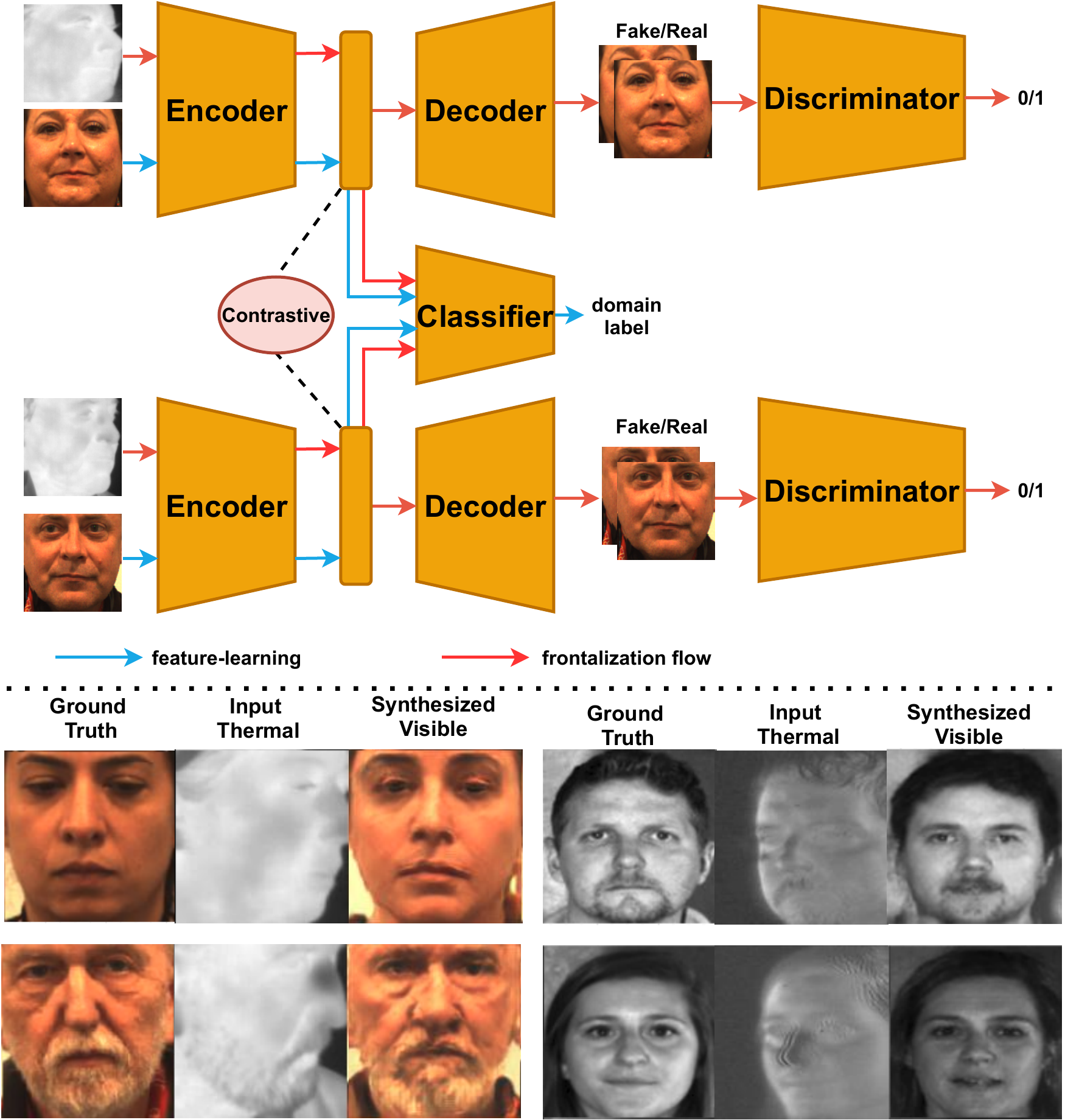}
	\caption{An overview of the proposed heterogeneous face frontalization method. {\color{red} Frontalization flow} aims to reconstruct a visible frontal face from a thermal profile face. {\color{myblue1} Feature-learning flow} aims to learn domain-agnostic features that imitate the domain discrepancy between input and output images. To enhance feature discrimination, the latent features are regularized by a contrastive constraint during training. }
	\label{fig:abstract}
	\vskip -10pt
\end{figure}

Face recognition is a challenging problem which has been actively researched over the past few decades.  In particular, deep   convolutional  neural  network (DCNN) based methods have shown  impressive performance improvements on various visible face recognition datasets.  Numerous DCNN-based models have been proposed in the literature which can address key challenges that include expression, illumination and pose variations, image resolution, aging, etc.   However, existing methods are specifically designed for recognizing face images that are collected in the visible spectrum.  In applications such as night-time surveillance, we are faced with \cmmnt{a scenario in which we have to identify} the scenario of identifying a face image acquired in the thermal domain by comparing it with a gallery of face images that are acquired in the visible domain. Existing DCNN-based visible face recognition methods will not perform well when directly applied to the problem of thermal to visible face recognition due to the significant distributional shift between the thermal and visible domains. In order to bridge this gap, various cross-domain face recognition algorithms have been proposed \cite{di2020multi,di2021multimodal,di2019facial,hu2015thermal,he2019adversarial,yu2019lamp,zhang2017generative,zhang2017tv,mallat2019cross,litvin2019novel,pereira2018domainunit,riggan2016estimation,Riggan2018thermal}.  In particular, synthesis-based methods have gained a lot of traction in recent years \cite{di2020multi}.  Given a thermal face image, the idea is to synthesize the corresponding face image in the visible domain.  Once the visible image is synthesized, any DCNN-based visible face recognition network can be leveraged for identification.

One of the main limitations of the existing synthesis-based models for cross-modal face recognition is that they do not perform well on thermal faces with large pose variations \cite{di2020multi}.  Face frontalization is an extensively studied problem in the computer vision and biometrics communities. Various methods have been developed for frontalization \cite{hassner2015effective, zhu2015high,sagonas2015robust,cao2018highfidelity,tran2017disentangled,hu2018pose,huang2017beyond,yin2017towards,yin2020dual,zhao2018towards,zhao2018-3daid,wei2020ffwm,zhang2018face}.  However, existing face frontalization methods' performance degrades significantly on thermal faces since they are specifically designed for frontalizing visible face images.  In order to deal with this problem, heterogeneous face frontalization methods are needed in which a model takes a thermal profile face image and generates a frontal visible face.  This is an extremely difficult problem due to the large domain as well as large pose discrepancies between the two modalities. Despite its applications in biometrics and surveillance, this problem is relatively unexplored in the literature.

We propose a domain agnostic learning-based generative adversarial network (DAL-GAN) \cite{goodfellow2014generative} with dual-path training architecture which can synthesize frontal views in the visible domain from thermal faces with pose variations.  Fig.~\ref{fig:abstract} gives a simplified overview of the proposed heterogeneous frontalization framework.  The model is trained on two flows: frontalization flow and feature-learning flow. The frontalization flow aims to synthesize a visible frontal face image from a thermal profile face image. The feature learning flow aims to learn domain-agnostic features that help in reconstructing better visible faces.  DAL-GAN consists of a generator with  an auxiliary classifier and two discriminators which capture both local and global texture discriminations for better  synthesis.  A contrastive constraint is enforced in the latent space of the generator with the help of a dual-path training strategy, which improves the feature vector's discrimination.  Finally, a multi-purpose loss function is utilized to guide the network in synthesizing identity-preserving cross-domain frontalization.   We conduct extensive experiments to demonstrate that DAL-GAN can generate better quality frontal views compared to the other baseline methods. Fig.~\ref{fig:abstract} shows sample outputs from the proposed network.

In summary, this paper makes the following contributions.
\begin{itemize}[noitemsep]
	
	\item We propose a cross-spectrum face frontalization model via learning domain-agnostic features. To the best of our knowledge, this is one of the first models for cross-spectrum face frontalization.
	
	\item We introduce a dual-path network architecture for learning  discriminative and domain-agnostic features. Features are obtained based on both a gradient reversal layer and the contrastive learning strategy.

	\item Extensive experiments and ablation studies are conducted to demonstrate the effectiveness of the proposed face frontalization method.  
	
\end{itemize}

\begin{figure*}[t!]
	\centering
	\includegraphics[width=0.85\linewidth]{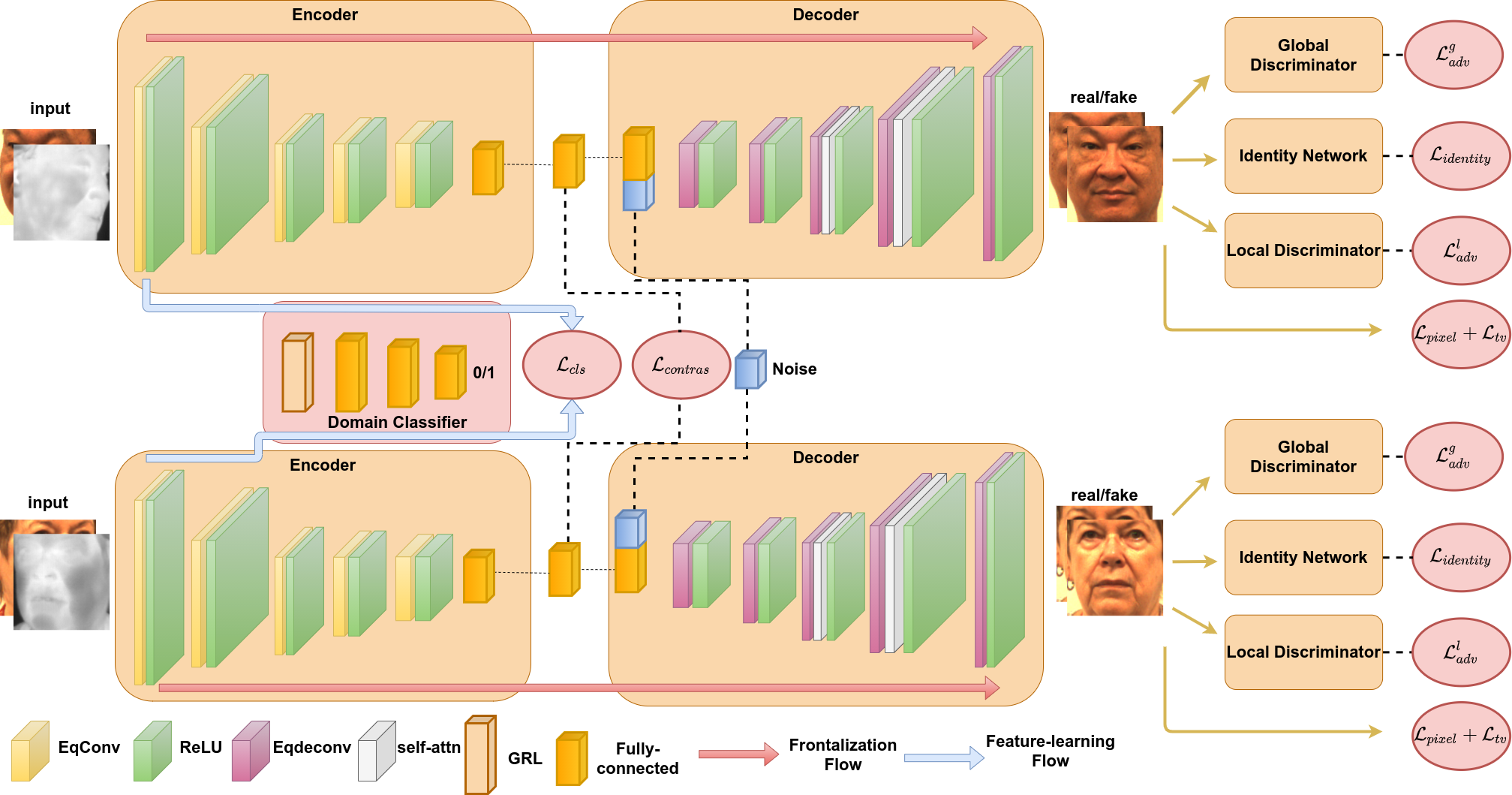}
	\caption{Illustration of the proposed dual-path architecture. Two weight-shared identical generators  are employed in each path. Both the face \textbf{frontalization flow} and domain-agnostic \textbf{feature-learning flow} are implemented during training. For frontalization, a proper combination of multiple losses are utilized which contain the multi-scale pixel loss $\mathcal{L}_{pixel}$, identity loss $\mathcal{L}_{id}$, global and local adversarial losses $\mathcal{L}^{g}_{adv}$ and $\mathcal{L}^{l}_{adv}$, total variation loss $\mathcal{L}_{tv}$ as well as contrastive loss $\mathcal{L}_{contras}$. Additionally, another domain classifier network is utilized for learning domain-agnostic features which are optimized by the classification loss $\mathcal{L}_{cls}$ with a gradient reversal layer (GRL). }
	\label{fig:architecturefrontal}
	\vskip -10pt
\end{figure*}

\section{Related Work}
\noindent \textbf{Face Frontalization:} Recent face frontalization methods utilize either 2D/3D warping \cite{hassner2015effective, zhu2015high,jeni2016person,zhu2015high,zhao2018-3daid}, stochastic modeling \cite{sagonas2015robust,booth20163d,booth2018large} or the generative models \cite{cao2018highfidelity,tran2017disentangled,hu2018pose,huang2017beyond,yin2017towards,yin2020dual,zhao2018towards,wei2020ffwm,zhang2018face,jiang2021geometrically}. For instance, Hassner \etal \cite{hassner2015effective} proposed a single unmodified 3D surface model for frontalization. Sagonas \etal \cite{sagonas2015robust} proposed a model that jointly performs frontalization and landmark localization by solving a low-rank optimization. 
Rui \etal \cite{huang2017beyond} developed TP-GAN with a two-path architecture for capturing both the global and local contextual information. Tran \etal \cite{tran2017disentangled} introduced the DR-GAN model which frontalizes face images via disentangled representations.  Hu \etal \cite{hu2018pose} introduced CAPG-GAN to synthesize  frontal face images guided by the target landmarks. Zhao \etal \cite{zhao2018towards} proposed the PIM model based on the domain adaptation strategy for pose invariant face recognition. Cao \etal \cite{cao2018highfidelity} proposed the HF-PIM model which includes dense correspondence field estimation and facial texture map recovery. Similarly, Zhang \etal \cite{zhang2018face} developed the A3F-CNN model with the appearance flow constrain. Li \etal \cite{li2019m2fpa} proposed a frontalization model using a series of discriminators optimized by segmented face images. Yin \etal \cite{yin2020dual} proposed a self-attention-based generator is to integrate local features with their long-range dependencies for obtaining better frontalized faces.  Wei \etal \cite{wei2020ffwm} proposed the FFWM model to overcome the illumination issue via flow-based feature warping.

\noindent \textbf{Thermal-to-visible Synthesis:} Various  approaches have been proposed in the literature for thermal-to-visible face synthesis and matching \cite{di2020multi,hu2015thermal,he2019adversarial,yu2019lamp,zhang2017generative,zhang2017tv,mallat2019cross,litvin2019novel,pereira2018domainunit,riggan2016estimation,Riggan2018thermal,di2019polarimetric,poster2021large,immidisetti2021simultaneous}. For instance, Hu \etal \cite{hu2015thermal} developed a model based on partial least squares for heterogeneous face recognition. Riggan \etal \cite{riggan2016estimation} combined feature estimation and image reconstruction steps for matching heterogeneous faces.  Zhang \etal \cite{zhang2017generative} introduced the GAN-VFS model for  cross-spectrum synthesis. Another TV-GAN model is introduced by Zhang \etal \cite{zhang2017tv} which leverages the identity information via a classification loss. Riggan \etal \cite{Riggan2018thermal} enhanced the image quality of the synthesized images by leveraging both global
and local facial regions. Moreover, some models \cite{zhang2018synthesis,pereira2018domainunit} synthesize high-quality images by leveraging complementary information in multiple modalities. Recently, Mallat \etal \cite{mallat2019cross} proposed a cascaded refinement network (CRN) model to synthesize visible images using a contextual loss. Litvin \etal \cite{litvin2019novel} developed the FusionNet architecture  to overcome the issue of model over-fitting on the RGB-T images \cite{Nikisins2014RGBDT}. Very recently, Di \etal introduced Multi-AP-GAN \cite{di2020multi} and AP-GAN \cite{di2018polarimetric} models for improving the quality of the synthesized visible faces using facial attributes.

Note that the proposed model is different from the models discussed above. In our approach, a feature normalization-based generative adversarial network  (GAN) is utilized to learn frontalization mapping. The generator is also optimized by learning domain-agnostic features through a gradient reversal layer (GRL) \cite{ganin2015unsupervised}. However, we find that simply learning domain-agnostic features is not enough for cross-spectrum frontalization, and learning discriminative features is also important. Therefore, semi-supervised contrastive learning is also implemented via a dual-path training strategy.

\section{Proposed Method}
Given a thermal face image $\mathbf{x}$ with a significant pose variation, our objective is to synthesize the corresponding frontal face image $\hat{\mathbf{y}}$ in the visible domain.  The generated frontal face image should be photo-realistic and identity-preserving.  In order to address this cross-spectrum face frontalization problem,  we propose a dual-path network architecture which learns domain-agnostic features via a contrastive learning strategy.  In what follows, we present the proposed network in detail.

\subsection{Networks Architecture}
Fig.~\ref{fig:architecturefrontal} gives an overview of the proposed dual-path architecture.  Each path contains a generator $F$ and two discriminators $D^{g}, D^{l}$. The two generators share weights.  Another domain classifier $C$ is utilized for learning domain-agnostic features. More details about the network architecture can be found in the supplementary file. 

\subsubsection{Generator} Each generator $F$ consists of an encoder-decoder structure where the encoder $E$ aims to extract domain-agnostic features from the input thermal face  while the decoder $G$ aims to reconstruct a frontal face image in the visible domain. In this work, we adopt the generator architecture from DA-GAN \cite{yin2020dual}  with the following modifications.

We remove all batch normalization layers in DA-GAN because they were originally introduced to eliminate the covariate shift. However,  recent studies have shown that   covariate shift often does not exist in GANs \cite{karras2018progressive,karras2019style,karras2020analyzing}. Instead we use feature vector equalization  to prevent the escalation of feature magnitudes.   Feature  equalization also helps the network converge smoothly during training. Given the original feature vector $a_{m,n}$ at pixel location $(m,n)$, the equalization is defined as follows
\setlength{\belowdisplayskip}{0pt} \setlength{\belowdisplayshortskip}{0pt}
\setlength{\abovedisplayskip}{0pt} \setlength{\abovedisplayshortskip}{0pt}

\begin{equation}\label{eq:feature equalization}
	\mathbf{b}_{m,n} = \mathbf{a}_{m,n}/ \sqrt{\frac{1}{N}\sum_{i=0}^{N-1}\mathbf{a}^{i}_{m,n} + \epsilon},
\end{equation}
 where $\mathbf{b}_{m,n}$ is the normalized vector and $N$ is the total number of features. We set $\epsilon$ equal to $ 10^{-8}$.  In our experiments, we find that feature equalization also allows us to use a higher initial learning rate which in turn helps in accelerating training.

\subsubsection{Discriminator} 
\begin{figure}[htp!]
	\centering
	\includegraphics[width=0.7\linewidth]{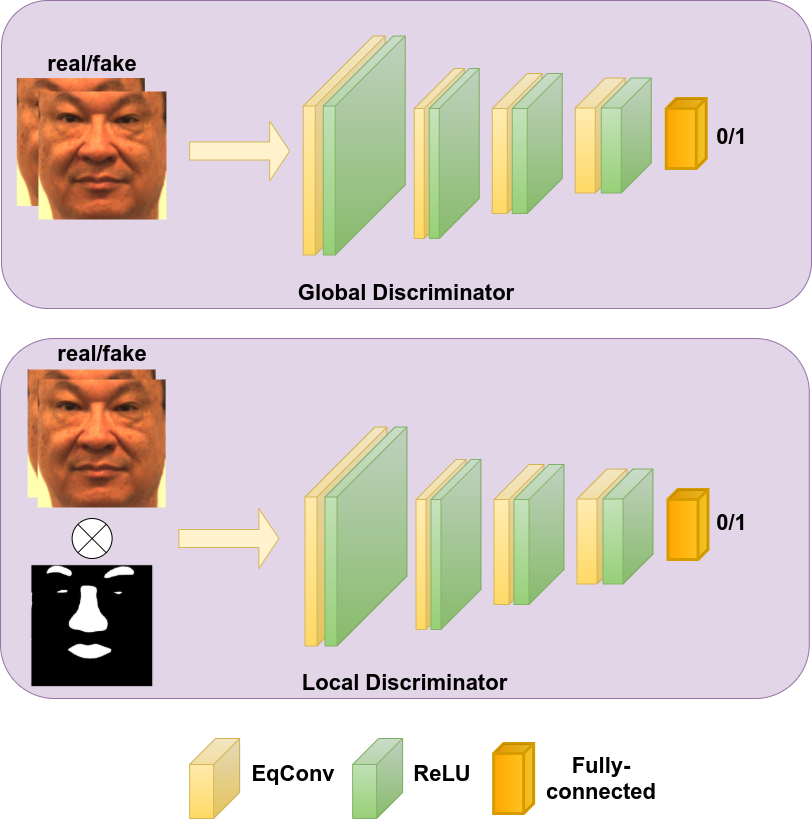}
	\caption{Illustration of the global and local discriminators.}
	\label{fig:driscriminators}
	\vskip -10pt
\end{figure}

A single discriminator may not be able to capture both global and local facial textures \cite{huang2017beyond,zhao2018towards}. Therefore, we employ two identical discriminators ($D_{l}, D_{g}$) to learn global and local discrimination respectively, as shown in Fig.~\ref{fig:driscriminators}. In particular, the local components of a frontal face $\mathbf{\hat{y}}_{f}$ are extracted by a predefined off-the-shelf model $F_{m}$  \cite{liu2015multi}. In this work, the key components (eyes, nose, lips, brow) are extracted to learn the local facial structure and the entire face image is used for learning the global texture. Mathematically, we define a mask $M$ to extract the key components of a ground-truth visible image $\mathbf{y}$ as: \[ \mathbf{M} = F_{m}(\mathbf{y}). \]  With the help of this mask, we can obtain local regions of  a real/fake image by the element-wise product $\odot$ between the mask and the real/fake image. The local/global discriminators are optimized by the following  loss function $\mathcal{L}_{adv}^{l}$/$\mathcal{L}_{adv}^{g}$ separately
\begin{equation}\label{adversarial loss}
	\begin{split}
	\mathcal{L}_{adv}^{g} =  \mathbb{E}[D_{g}(\mathbf{y})] - \mathbb{E}[ (D_{g}(\mathbf{\hat{y}}))] - \\ 
	\lambda_{gp} \mathbb{E}[(\|\triangledown_{\mathbf{y}^{\star}} D_{g}(\mathbf{y}^{\star}) \|_{2} -1)^{2}], \\	
	\mathcal{L}_{adv}^{l} =  \mathbb{E}[D_{l}(\mathbf{M} \odot \mathbf{y})] - \mathbb{E}[ (D_{l}(\mathbf{M} \odot \mathbf{\hat{y}}))] - \\ 
	\lambda_{gp} \mathbb{E}[(\|\triangledown_{\mathbf{M} \odot \mathbf{y}^{\star}} D_{l}(\mathbf{M} \odot \mathbf{y}^{\star}) \|_{2} -1)^{2}], \\
	\mathcal{L}_{adv} = \mathcal{L}^{g}_{adv} + \lambda_{l} \cdot \mathcal{L}^{l}_{adv}.
	\end{split}
\end{equation}
Here, $\mathbf{y}^{\star}_{f}$ is sampled uniformly along a straight line between a pair of real image $\mathbf{y}_{f}$ and the generated image $\mathbf{\hat{y}}_{f}$ \cite{Ishaan2017wgangp}. The
discriminators attempts to maximize this objective, while the generator attempts to minimize it. We set $\lambda_{gp} = 10,  \lambda_{l} = 0.1$ in our experiments.

\subsection{Objective Function}
The loss function we use to train the proposed network consists of a combination of contrastive loss, gradient reversal layer-based domain adaptive loss and frontalization loss.\\

\noindent {\bf{Frontalization Loss:}}  To capture texture information between the real and fake images at multiple scalese, we utilize a multi-scale pixel loss as follows
\begin{equation}\label{eq:pixel loss}
	\begin{split}
		\mathcal{L}_{pixel} = \sum_{s=1}^{3}\|\mathbf{\hat{y}}^{s} -\mathbf{y}^{s}\|_{1},
	\end{split}
\end{equation}
where $s$ corresponds to the resolution scale.  In our work, we utilize three different resolution scales: $32\times 32$, $64\times 64$ and $128 \times 128$.

Identity preserving synthesis is important for heterogeneous face recognition. To achieve this, we utilize the identity loss as follows 
\begin{equation}\label{eq:identity loss}
	\begin{split}
		\mathcal{L}_{id} = \|F_{v}(\mathbf{\hat{y}}) -F_{v}(\mathbf{y})\|_{2},
	\end{split}
\end{equation}
where $F_{v}$ is the pre-trained VGGFace \cite{parkhi2015deep} model which is used to extract  features. By minimizing the feature distance between the real and fake images, the generator is optimized to synthesize identity-preserving images.

In order to reduce the artifacts in the synthetic images, a total variation regularization $\mathcal{L}_{tv}$ is also applied as that in \cite{johnson2016perceptual}. Hence, the total loss for the frontalization flow is defined as follows
\begin{equation}\label{eq: frontalization loss}
	\mathcal{L}_{front} = \mathcal{L}_{pixel} + \lambda_{I} \cdot \mathcal{L}_{id} + \lambda_{adv} \cdot \mathcal{L}_{adv} + \lambda_{tv} \cdot \mathcal{L}_{tv}, \qquad	\\	
\end{equation}
 where $\lambda_{I}, \lambda_{adv},  \lambda_{tv}$ are  parameters.\\

\noindent {\bf{Domain Adaptive Loss:}}  To learn domain-agnostic features, one domain classifier network $C$ with a gradient reversal layer \cite{ganin2015unsupervised} is employed in our network. The domain classifier network is a simple 3-layer multi-linear perceptron  (MLP) neural network with the same equalization as in Eq~\eqref{eq:feature equalization}. Given both thermal and visible face images, the classifier aims to correctly estimate which spectrum the input image belongs to. When back-propagating, the gradient reversal layer flips the gradients which leads to domain-agnostic features in the encoder. The network parameters $\theta_{E}$ corresponding to the encoder network are updated as follows
\begin{equation}\label{gradiennt}
	\small
	\begin{split}
	\mathcal{L}_{cls} = \frac{1}{N} \sum_{i=0}^{N-1} - \mathbf{k}_{i} \log(C(E(\mathbf{x}_{i}))) - (1-\mathbf{k}_{i}) \log(1-C(E(\mathbf{x}_{i}))) \\
	\theta_{E}' = \theta_{E} - \mu \left(\frac{\partial \mathcal{L}_{front}}{\partial \theta_{E}} - \lambda_{grl} \frac{\partial \mathcal{L}_{cls}}{\partial \theta_{E}}\right), \qquad \qquad \qquad
	\end{split}
\end{equation}
where $\mathbf{x}_{i}$ is the $i$-th input sample, $\mu$ is the learning rate, $\theta_{E}'$ and $\theta_{E}$ are the updated and original parameters, respectively. $\mathbf{k}_{i}$ is the domain index $[0,1]$. $\lambda_{grl}$ is set equal to $0.01$ in our experiments.\\

\noindent {\bf{Contrastive Loss:}}
Finally, the contrasitive loss \cite{chopra2005learning} is used to enhance the discriminability of the latent features.  It is defined as follows
 \begin{equation}\label{eq:latent contrasitive}
	\begin{split}
		\mathcal{L}_{Contras} = \mathbf{l} \cdot \|E(\mathbf{x}^{1}) -E(\mathbf{x}^{2})\|_{2} + \\  (1-\mathbf{l}) \cdot  \max (0, m-\|E(\mathbf{x}^{1}) -E(\mathbf{x}^{2})\|_{2}),
	\end{split}	
\end{equation}
where $\mathbf{x}^{1}, \mathbf{x}^{2}$ are two profile face images input to the dual-path architecture. The label vector $\mathbf{l}$ indicates whether they contain the same ($\mathbf{l}=1$) or different ($\mathbf{l}=0$) identity. The hyper-parameter $m=1.2$ indicates the margin.\\

\noindent {\bf{Total Objective: }}  In order to reduce the artifacts in the synthetic images, a total variation regularization $\mathcal{L}_{tv}$ is also applied as that in \cite{johnson2016perceptual}. The overall loss function used to train the proposed heterogeneous frontalization network is as follows
\begin{equation}\label{total objective}
	\begin{split}
	\mathcal{L} =  \mathcal{L}_{front} + \lambda_{C} \cdot \mathcal{L}_{contras}  + \lambda_{cls} \cdot \mathcal{L}_{cls}, \qquad \qquad \quad \\
	\end{split}
\end{equation} where $\lambda_{con}, \lambda_{cls}$ are regularization parameters.

\section{Experiments}

\begin{figure}[htp!]
	\centering
	\includegraphics[width=0.95\linewidth]{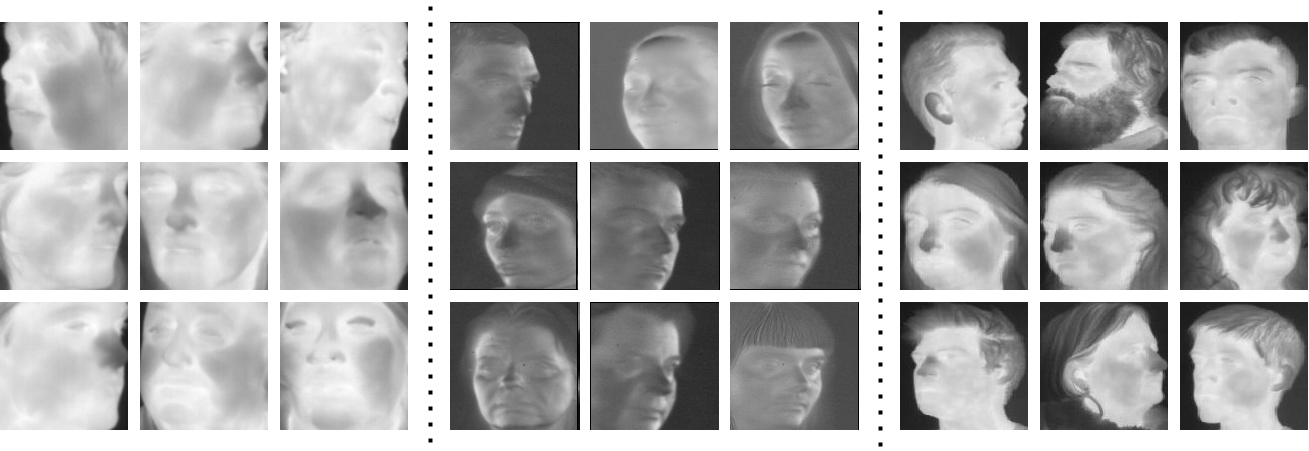} \\
	\raggedright \hskip 12pt ARL-VTF \cite{dominick2021time} \hskip 10pt ARL-MMFD \cite{di2020multi} \hskip 12pt TUFTS \cite{panetta2018comprehensive}
	\caption{Input profile thermal images sampled from three datasets respectively.}
	\label{fig:dataset-samples}
\end{figure}

\noindent {\bf{Datasets:}} We evaluate the proposed heterogeneous face frontalization network on three publicly available datasets:  DEVCOM Army Research Laboratory Visible-Thermal Face Dataset (ARL-VTF) \cite{dominick2021time}, ARL Multimodal Face Database (ARL-MMFD) \cite{di2020multi,hu2016polarimetric,zhang2018synthesis} and TUFTS Face \cite{panetta2018comprehensive}.  Sample images from these datasets are shown in Fig.~\ref{fig:dataset-samples}.\\

\noindent {\bf{\emph{ARL-VTF Dataset: }}}  The ARL-VTF dataset \cite{dominick2021time} consists of 500,000 images from 395 subjects (295 for training, 100 for testing).  This dataset contains a large collection of paired visible and thermal faces. Variations in baseline, expressions, pose, and eye-wear are included.   We first crop the images based on the given ground-truth bounding-box annotations for conducting experiments.
The proposed model is trained on this dataset by the predefined development and testing splits \cite{dominick2021time}. The final experimental results are based on the average of  the predefined 5 splits.\\

\noindent {\bf{\emph{ARL-MMFD Dataset: }}}  Additionally, we evaluate the proposed model on Volume \RNum{3} of the ARL-MMFD dataset \cite{di2020multi}. This dataset  was collected by ARL cross 11 sessions over 6 days. It contains 5419 paired polarimetric thermal and visible images from 121 subjects with significant variations in expression, illuminations, pose, glasses, etc. We select the polarimetric thermal profile and visible frontal face image pairs corresponding to both neutral and expressive faces for conducting experiments on this dataset.  In particular, we randomly select images from 90 identities for training and the images from the remaining identities for testing. So, there are no overlapping images between training and testing sets. We use the original aligned visible and polarimetric thermal images for training and testing without any preprocessing. The final experimental results are based on the average of  five random  splits.\\

\noindent {\bf{\emph{TUFTS Face Dataset: }}}  Finally, the proposed model is also evaluated on a recently published TUFTS Face Dataset \cite{panetta2018comprehensive}.  TUFTS is a multi-modal dataset which contains more than 10,000 images from 113 individuals from more than 15 countries. For conducting experiments, we only select thermal and visible images from this multi-modal dataset.  Hence, there are more than 1000 images from over 100 subjects with different pose and expression variations. This dataset is very challenging due to a large number of pose and expression variations and only a few images per variation are available in the dataset. Images from randomly selected  89 individuals are used for training and images from the remaining 23 subjects are used for testing. The raw images are used to train and test without any pre-processing. In particular, profile images in the thermal domain and the frontal images (both neural and expression) in the visible domain are utilized for training the models.\\ 


\noindent \textbf{Implementation Details:} All the images in this work are resized to $ 128 \times 128$ and the image intensity is scaled into [0, 1]. The features from VGGFace \cite{Parkhi15DeepFace} average pooling layer are extracted from the synthesized visible image and the similarity for verification is calculated based on the cosine distance. In all the experiments, the hyper-parameters are set as $\lambda_{I}=10, \lambda_{adv}=1, \lambda_{C}=0.01, \lambda_{tv}=1e^{-4}$. The learning rate is initially set as 0.01 and the batch size is 8. We train our model 10, 100, and 400 epochs on ARL-VTF, ARL-MMFD and TUFTS datasets, respectively. 


We compare the proposed method with the following state-of-the-art facial frontalization methods: TP-GAN  \cite{huang2017beyond}; PIM \cite{zhao2018towards}; M2FPA \cite{li2019m2fpa}; DA-GAN \cite{yin2020dual}. Besides, we add pix2pix \cite{pix2pix2017} as a baseline, which is an image-to-image translation method between two domains. We followed the same network settings as mentioned in the  original papers and tried our best to finetune the training parameters. Note that TP-GAN \cite{huang2017beyond} and PIM \cite{zhao2018towards} require landmarks on input profile thermal face images. We manually label the landmarks on thermal faces in the  TUFTS Face Dataset \cite{panetta2018comprehensive} and use officially provided landmarks in the ARL-VTF dataset \cite{dominick2021time}.  We estimate the landmarks by MTCNN \cite{zhang2016joint} on the images in the  ARL-MMFD dataset \cite{di2020multi}.

\subsection{Experimental Results}


\noindent {\bf{ARL Visible-Thermal Face Dataset:}} 
The evaluation is based on the following two protocols: (1) Gallery G\_VB0- to Probe P\_PTP0. (2) Gallery G\_VB0- to Probe P\_PTP-. The images in Gallery G\_VB0- are the facial baseline images in the visible domain without eye-glasses occlusion. The images in Probe P\_PTP0 are the facial profile images in the thermal domain without eye-glasses occlusion. The images in Probe P\_PTP- are the facial profile images in the thermal domain with eye-glasses token-off.

\begin{figure}
	\centering
	\includegraphics[width=0.95\linewidth]{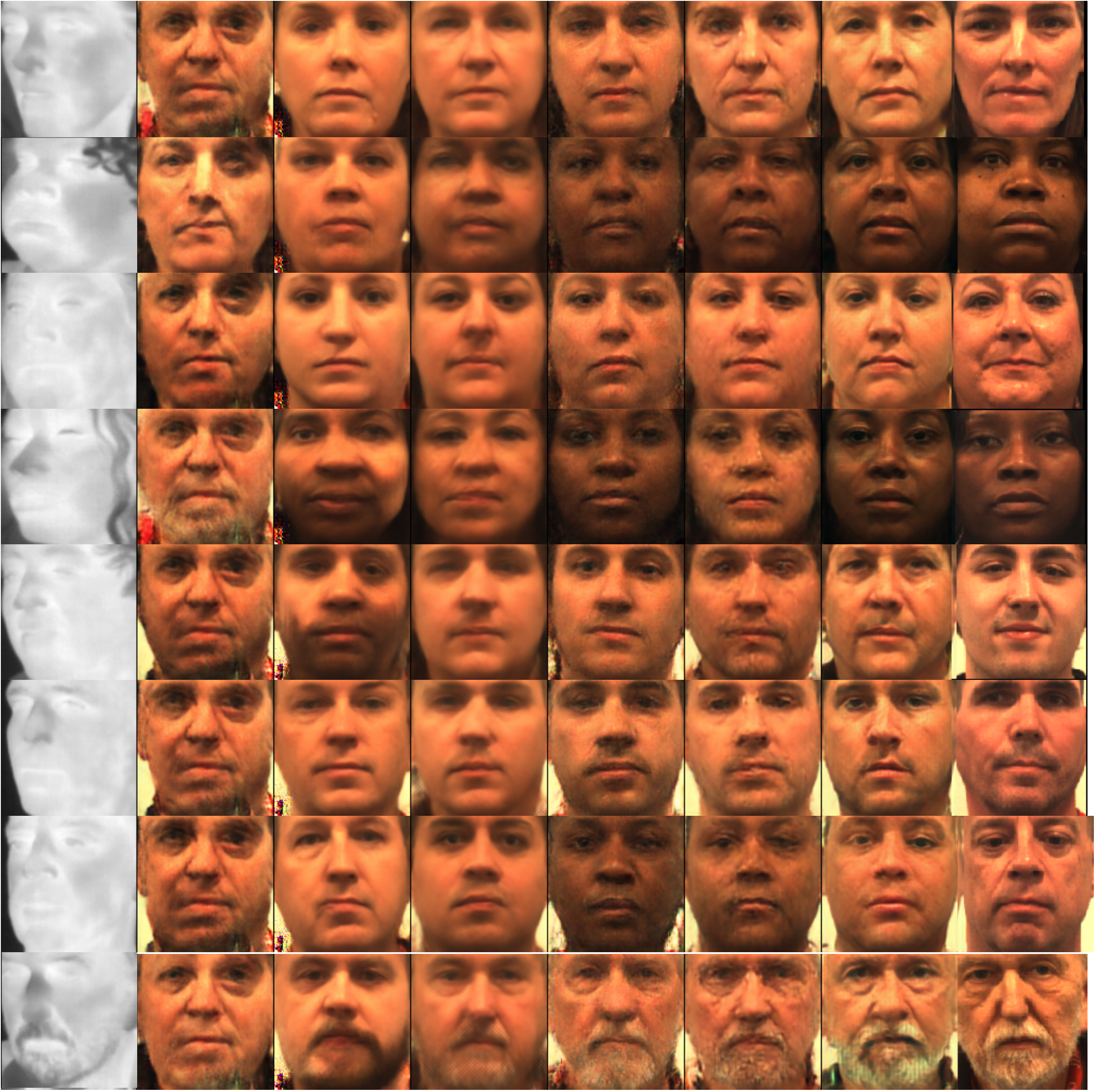} \\
	\tiny
	\raggedright \hskip 15pt Input \hskip 8pt Pix2Pix\cite{pix2pix2017} \hskip 2pt TP-GAN \cite{huang2017beyond} \hskip 5pt PIM\cite{zhao2018towards} \hskip 7pt M2FPA \cite{li2019m2fpa} \hskip 1pt DA-GAN\cite{yin2020dual} \hskip 9pt Ours \hskip 12pt Reference
	\caption{Cross--domain face frontalization comparison on the ARL-VTF \cite{dominick2021time} dataset.}
	\label{fig:odinbaselinecomparison}
	\vskip -10pt
\end{figure}

We evaluate our model with the other baseline methods both qualitatively and quantitatively. Visual frontalization results are shown in Fig.~\ref{fig:odinbaselinecomparison}.
As can be seen from this figure, the proposed method is able to synthesize more photo-realistic and identity-preserving images compared with the other methods. Other methods fail to synthesize high-quality  and identity-preserving images.  Additionally, Table~\ref{tb: comparison_table} quantitatively compares the verification performance of different methods. Our method achieves around 2\% and 5\% improvements on both the AUC and EER scores in two protocols, respectively when compared with the recent DA-GAN \cite{yin2020dual}. Additionally,
the proposed model also achieves $2\% \sim 5\%$ improvements on the True Accept Rate (TAR) at False Accept Rates (FAR)  1\% and 5\%.\\

\begin{table*}[t]
	\caption{Verification performance comparison on the ARL-VTF dataset \cite{dominick2021time}.}
	\centering
	\resizebox{1.8\columnwidth}{!}{%
		\begin{tabular}{|c|c|c|c|c|c|c|c|c|c|}
			\hline & Probe & \multicolumn{4}{|c|}{00\_pose} & \multicolumn{4}{|c|}{10\_pose} \\
			\hline Gallery & Metric & AUC & EER &  FAR=1\% & FAR=5\% & AUC & EER & FAR=1\% & FAR=5\% \\ 
			\hline
			\multicolumn{1}{|c|}{\multirow{5}{*}{Gallery 0010}} 
			& Raw & 54.88 & 46.38 & 2.16  & 8.33 & 56.10 & 45.98 & 1.06  & 8.53 \\
			& Pix2Pix \cite{pix2pix2017} & 5.04 & 46.84 & 2.17  & 8.90 & 57.95  & 44.84  & 1.79  & 14.47 \\ 
			& TP-GAN \cite{huang2017beyond} & 64.21 & 40.12  & 3.31  & 12.11 & 67.41 & 36.78  & 3.89  & 13.84 \\
			& PIM \cite{zhao2018towards} & 68.69 & 36.58 & 5.43 & 16.56 & 73.31 & 32.70 & 5.96 & 20.42 \\
			& M2FPA \cite{li2019m2fpa} & 74.99 & 32.33 & 5.73 & 20.26 & 76.99 & 29.84 & 9.51 & 23.35 \\
			& DA-GAN \cite{yin2020dual} & 75.58 &  31.18 & 6.85  & 22.23 & 75.76 & 30.69  & 8.40  & 23.62 \\ 
			& Ours  & 77.48 &  29.08 & 8.20  &  25.89 &  82.18 & 25.11 &  10.82 & 30.64 \\
			\hline 	
		\end{tabular}
	}	
	\label{tb: comparison_table}
	\vskip -2pt
\end{table*}

\begin{table*}[t]
	\caption{Verification performance comparison corresponding to the ablation study.}
	\centering
	\resizebox{1.8\columnwidth}{!}{%
		\begin{tabular}{|c|c|c|c|c|c|c|c|c|c|}
			\hline & Probe & \multicolumn{4}{|c|}{00\_pose} & \multicolumn{4}{|c|}{10\_pose} \\
			\hline Gallery & Metric & AUC & EER &  FAR=1\% & FAR=5\% & AUC & EER & FAR=1\% & FAR=5\% \\ 
			\hline
			\multicolumn{1}{|c|}{\multirow{5}{*}{Gallery 0010}} 
			& Baseline & 56.21 & 45.99 &  2.07  & 9.04  & 59.20 & 43.13 & 2.73 & 9.71   \\
			& w/ Multi-scale $\mathcal{L}_{1}$ & 58.07 &  44.82 & 2.13 & 10.27  & 62.01 &  40.91 & 3.84  & 11.06  \\
			& w/ $\mathcal{L}_{id}$ & 70.26 & 34.26  & 4.90  & 16.26 & 76.56 & 29.67 & 6.70 & 23.84 \\	
			& w/ self-attn & 71.99 & 34.05  & 5.22  & 19.72 &  76.60 & 29.94  & 6.62  & 24.00 \\
			& w/ $\mathcal{L}^{l}_{adv}$ & 72.58 & 33.54  & 6.21  & 20.07 &  76.58 & 28.44  & 6.66  & 24.18 \\
			& w/ Eq.\eqref{eq:feature equalization}  & 77.13 & 29.16  & 6.41  & 21.35 &  80.05 & 26.26  & 7.11  & 29.40 \\
			& w/ $\mathcal{L}_{cls}$  & 77.18 & 29.07  & 7.43  & 22.20 &  80.56 & 26.46  & 9.24  & 29.90 \\	
			& w/ $\mathcal{L}_{contras}$ (ours) & 77.48 &  29.08 & 8.20  &  25.89  & 82.18 & 25.11 &  10.82 & 30.64 \\
			\hline	
		\end{tabular}
	}	
	\label{tb: component study table}
	\vskip -2pt
\end{table*}

\begin{figure}
	\centering
	\includegraphics[width=0.95\linewidth]{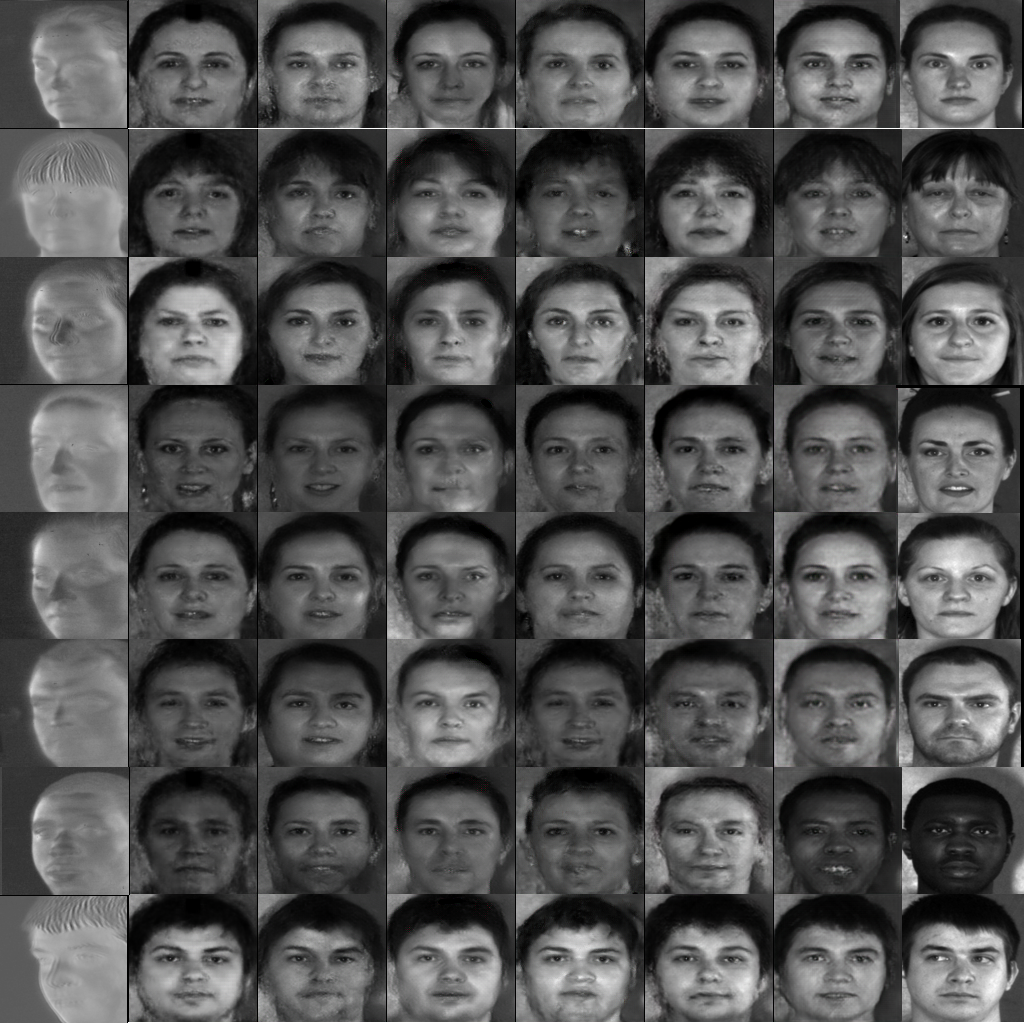} \\
	\tiny
	\raggedright \hskip 15pt Input \hskip 8pt Pix2Pix\cite{pix2pix2017} \hskip 2pt TP-GAN \cite{huang2017beyond} \hskip 5pt PIM\cite{zhao2018towards} \hskip 7pt M2FPA \cite{li2019m2fpa} \hskip 2pt DA-GAN\cite{yin2020dual} \hskip 9pt Ours \hskip 12pt Reference
	\caption{Cross-domain face frontalization comparison on the ARL-MMFD \cite{di2020multi} dataset.}
	\label{fig:arlbaselinecomparison}
	\vskip -10pt
\end{figure}

\noindent {\bf{ARL Multimodal Face Database: }}
We evaluate the proposed model on Volume \RNum{3} of ARL-MMFD \cite{di2020multi}. Qualitative and quantitative results corresponding to this dataset are shown in Fig.~\ref{fig:arlbaselinecomparison} and Table~\ref{tb: arl_comparison_table}, respectively.  As can be seen from Table~\ref{tb: arl_comparison_table}, the proposed model surpasses the best performing baseline model by 2.4\% and 1.5\% in AUC and EER scores, respectively.  Furthermore, the proposed model surpasses 7.7\% and 1.3\% when compared with the DA-GAN \cite{yin2017towards} at FAR=1\% and FAR=5\%, respectively.

From Fig.~\ref{fig:arlbaselinecomparison}, we can see that our model is able to synthesize more photo-realistic images while preserving the identity better than the other frontalization models. For each subject in this dataset, the pose variations mainly cover from $-60^{\circ} \sim 60^{\circ}$, while the illumination variation is various. Hence, the method like PIM \cite{zhao2018towards} fails to synthesize photo-realistic images due to the illumination artifacts. In addition, our proposed model is trained based on contrastive loss which helps the network provide better results even on this smaller dataset.\\
 

\begin{table}[t]
	\caption{Verification performance comparison on the ARL-MMFD dataset \cite{di2020multi}.}
	\centering
	\resizebox{1\columnwidth}{!}{%
		\begin{tabular}{|c|c|c|c|c|}
			\hline Method  & AUC & EER &  FAR=1\% & FAR=5\%  \\ 
			\hline
			Raw &  64.12 & 40.51 & 4.21  & 17.74  \\
			Pix2Pix \cite{pix2pix2017} & 73.60 & 31.96 &  11.16 & 26.45  \\
			TP-GAN \cite{huang2017beyond}  & 76.15 & 30.89 & 6.26  & 19.03  \\
			PIM \cite{zhao2018towards}  & 80.89 & 26.86 & 9.83  & 27.11  \\
			M2FPA \cite{li2019m2fpa}  & 85.58  & 22.27 & 13.32  & 37.58  \\
			DA-GAN \cite{yin2020dual}  & 86.26 & 21.56 & 14.74  & 33.17 \\
			Ours   & 88.61 & 20.21  &  16.07  & 40.93  \\
			\hline	
		\end{tabular}
	}	
	\label{tb: arl_comparison_table}
	\vskip -10pt
\end{table}

\noindent {\bf{TUFTS Face Database:}}  Finally, the proposed model is also evaluated on the TUFTS Face dataset \cite{panetta2018comprehensive}. The performances are reported in quantitative as shown in Table~\ref{tb: tufts_comparison_table}. We can observe our proposed method surpasses the previous baselines and achieves around $3\%$ improvement in AUC and EER scores.

\begin{table}[t]
	\caption{Verification performance comparison on the TUFTS Face Database \cite{panetta2018comprehensive}.}
	\centering
	\resizebox{0.8\columnwidth}{!}{%
		\begin{tabular}{|c|c|c|c|c|}
			\hline Method  & AUC & EER &  FAR=1\% & FAR=5\%  \\ 
			\hline
			Raw & 67.55 & 37.88 & 5.11  & 16.11  \\
			Pix2Pix \cite{pix2pix2017} & 69.71 & 35.31 & 5.44  & 21.66  \\
			TP-GAN \cite{huang2017beyond}  & 70.93 & 35.32 & 6.46  & 18.77  \\
			PIM \cite{zhao2018towards}  & 72.84 & 34.10 & 8.77  & 21.00  \\
			M2FPA \cite{li2019m2fpa}  & 75.07  & 31.22 & 8.33  & 23.44  \\
			DA-GAN \cite{yin2020dual}  & 75.24 & 31.14 & 10.44  & 26.22 \\
			Ours   & 78.68 & 28.38  &  10.44  & 27.11  \\
			\hline	
		\end{tabular}
	}	
	\label{tb: tufts_comparison_table}
\end{table}

\subsection{Ablation Study}

\begin{figure}
	\centering
	\includegraphics[width=0.95\linewidth]{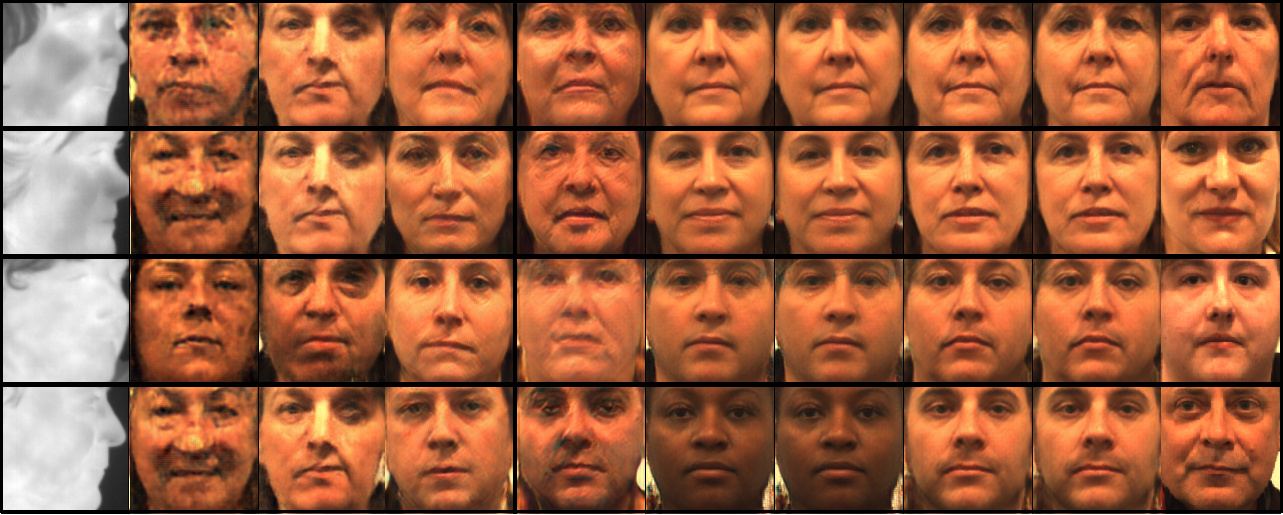}\\
	\tiny
	\raggedright \hskip 10pt Profile \hskip 8pt Baseline \hskip 5pt  w/ $\mathcal{L}_{1}$ \hskip 2pt  w/ $\mathcal{L}_{id}$ \hskip 2pt w/ self-attn \hskip 2pt w/ $\mathcal{L}^{l}_{adv}$ \hskip -2pt w/ Eq.~\eqref{eq:feature equalization} \hskip 2pt w/ $\mathcal{L}_{cls}$ \hskip 5pt  Ours \hskip 10pt  Frontal
	\caption{Results corresponding to the ablation study.}
	\label{fig:ablationstudy}
	\vskip -10pt
\end{figure}

In this section, we analyze how each part of the proposed model contributes to the final performance. We choose the global UNet \cite{huang2017beyond} generator and the discriminator $D_{g}$ as the baseline model. In particular, we analyze the contribution of each loss function and the equalization in Eq~\eqref{eq:feature equalization}. We conduct these ablation experiments on the ARL-VTF dataset and  show the results both qualitatively and quantitatively in Fig.~\ref{fig:ablationstudy} and Table~\ref{tb: component study table}, respectively. 

As shown in Table~\ref{tb: component study table}, the quantitative verification results are improved by consecutively adding different  components. Fig.~\ref{fig:ablationstudy} shows the corresponding synthesized samples. Based on this ablation study, we can see that  $\mathcal{L}_{id}$ significantly preserves the identity in the synthesized image.   Feature equalization Eq.~\eqref{eq:feature equalization} significantly improves the image quality. Finally, $\mathcal{L}_{cls}$ and $\mathcal{L}_{contras}$  improve the verification performance. 

\subsection{Pose Invariant Representation}
\begin{figure}
	\centering
	\includegraphics[width=0.9\linewidth]{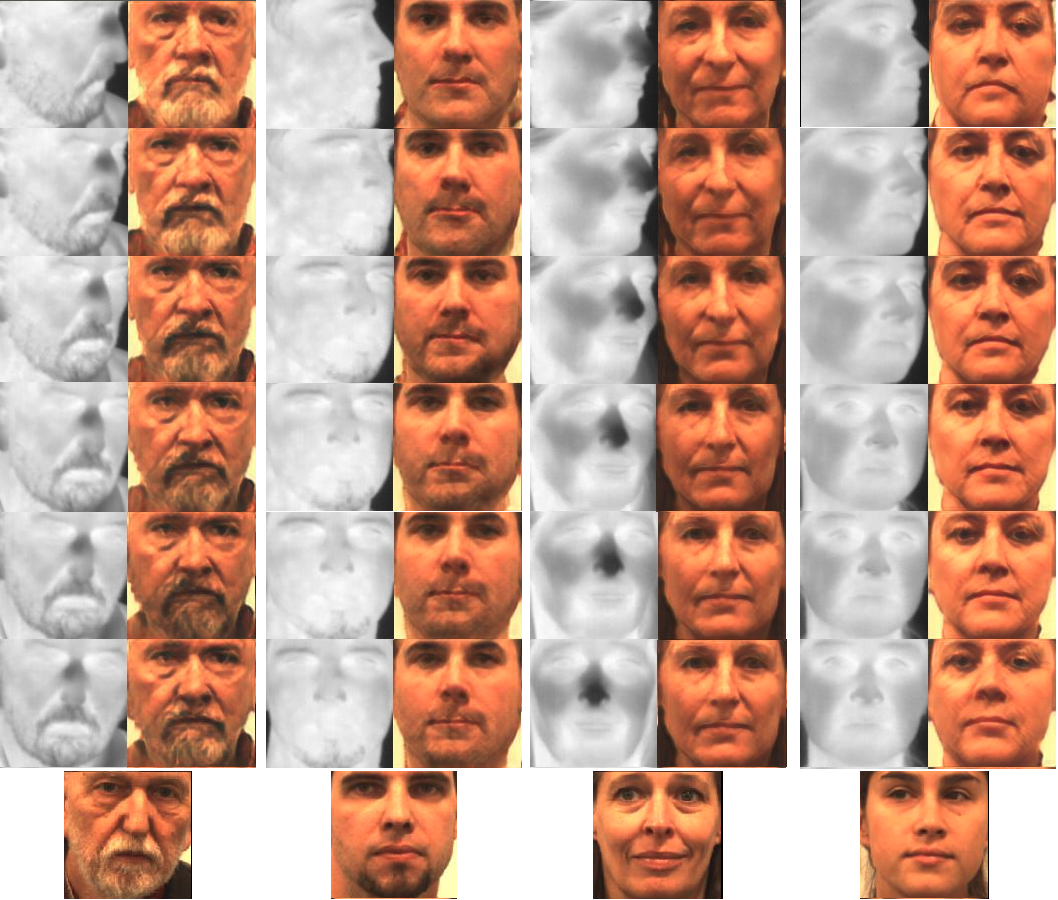}
	\caption{Synthesized frontal images  corresponding to a range of yaw poses on the ARL-VTF dataset.}
	\label{fig:odinposedegree}
	\vskip -10pt
\end{figure}

\begin{figure}
	\centering
	\includegraphics[width=0.9\linewidth]{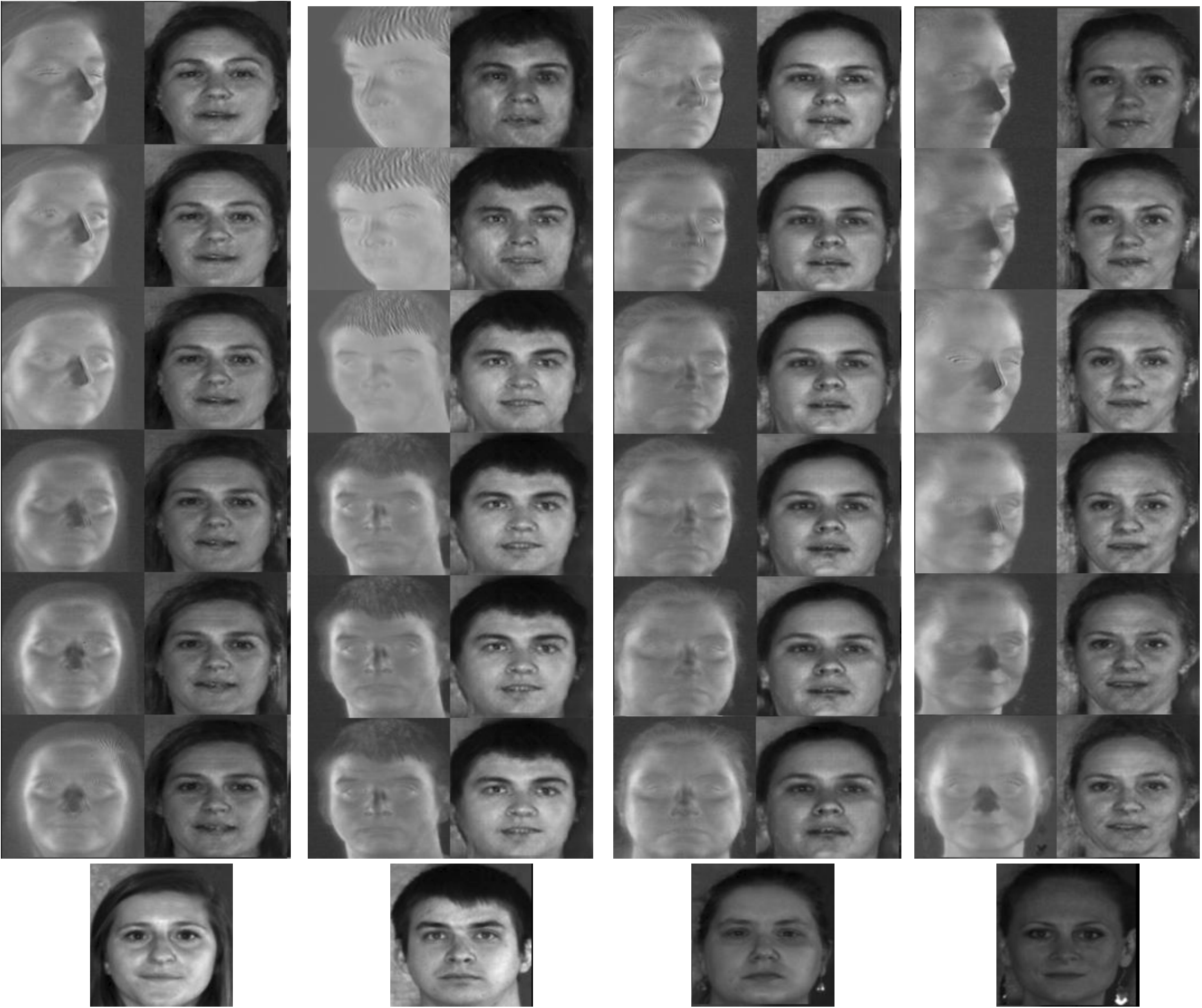}
	\caption{Synthesized frontal images  corresponding to a range of yaw poses on the ARL-MMFD dataset.}
	\label{fig:arlposedegree}
	\vskip -10pt
\end{figure}

In Fig.~\ref{fig:arlposedegree}, we show the synthesized frontal faces from the proposed model with a range of yaw poses on the ARL-MMFD dataset. Yaw poses in this dataset mainly cover $0^{\circ} \sim 45^{\circ} $. Given arbitrary yaw pose polarimetric thermal images, we can observe that the synthesized visible frontal images maintain a good pose-invariant representation. Additionally, we also conduct this analysis on the ARL-VTF dataset as shown in Fig.~\ref{fig:odinposedegree}. From these two figures, we can see that the proposed model learns pose-invariant frontal representation in the visible domain when given arbitrary pose thermal input images. 

\section{Conclusion}

In this work, we proposed a novel heterogeneous face frontalization model which generates frontal visible faces from profile thermal faces. The generator contains a gradient reversal layer-based classifier for domain-agnostic feature learning and a pair of local and global discriminators for better synthesis. Additionally, another contrastive constraint is enforced by a dual-path training strategy for learning discriminative latent features. Quantitative and visual experiments conducted on three real thermal-visible datasets demonstrate the superiority of the proposed method when compared to other existing methods. Additionally, an ablation study was conducted to demonstrate the improvements obtained from different modules and loss functions.

{\small
\bibliographystyle{ieee}
\bibliography{HF_frontal_cameraready_xing}
}

\end{document}